\documentclass{article}
\usepackage{spconf,amsmath,graphicx,amssymb}
\usepackage[ruled]{algorithm2e}
\usepackage{bm}
\usepackage[OT1]{fontenc}

\usepackage{titlesec}
\usepackage{times}
\usepackage{soul}
\usepackage{url}
\usepackage[hidelinks]{hyperref}
\usepackage[utf8]{inputenc}
\usepackage{booktabs}
\usepackage{color}
\usepackage{stfloats}
\usepackage{multirow}
\usepackage{array}
\usepackage[skins]{tcolorbox}
\usepackage{cleveref}
\newcommand{\PreserveBackslash}[1]{\let\temp=\\#1\let\\=\temp}
\newcolumntype{C}[1]{>{\PreserveBackslash\centering}p{#1}}
\crefname{equation}{}{}

\long\def\comment#1{}

\title{\MakeLowercase{$t$-$k$-means}: A Robust and Stable \MakeLowercase{$k$-means} Variant}
\name{Yiming Li$^{1,\star}$\thanks{$^\star$ indicates equal contribution.}, Yang Zhang$^{1,\star}$\thanks{This work is supported by the National Science Foundation of China under Grant 61771273, the Natural Science Foundation of Zhejiang Province (LSY19A010002), and the R\&D Program of Shenzhen (JCYJ20180508152204044). Corresponding author: Shu-Tao Xia (email: xiast@sz.tsinghua.edu.cn). }, Qingtao Tang$^{1}$, Weipeng Huang$^{2}$, Yong Jiang$^{1,3}$, Shu-Tao Xia$^{1,3}$} 

            \address{$^{1}$Tsinghua Shenzhen International Graduate School, Tsinghua University, China\\
            $^{2}$ Insight Centre for Data Analytic, University College Dublin, Ireland\\
            $^{3}$ PCL Research Center of Networks and Communications, Peng Cheng Laboratory, China}

\begin{document}
\ninept
\maketitle

\begin{abstract}
$k$-means algorithm is one of the most classical clustering methods, which has been widely and successfully used in signal processing. However, due to the thin-tailed property of the Gaussian distribution, $k$-means algorithm suffers from relatively poor performance on the dataset containing heavy-tailed data or outliers. Besides, standard $k$-means algorithm also has relatively weak stability, $i.e.$ its results have a large variance, which reduces its credibility. In this paper, we propose a robust and stable $k$-means variant, dubbed the $t$-$k$-means, as well as its fast version to alleviate those problems. Theoretically, we derive the $t$-$k$-means and analyze its robustness and stability from the aspect of the loss function and the expression of the clustering center, respectively. Extensive experiments are also conducted, which verify the effectiveness and efficiency of the proposed method. The code for reproducing main results is available at \url{https://github.com/THUYimingLi/t-k-means}.
\end{abstract}


\section{Introduction}
Data clustering is a fundamental problem in data mining, which has been widely and successfully adopted in many fields \cite{ollikainen2018clustering,berberidis2019node,kiselev2019challenges}. In the past decade, many different clustering algorithms \cite{lloyd1982least,chang2017deep,peng2018structured} have been proposed by researchers from different fields. Among all these methods, Lloyd's $k$-means algorithm \cite{lloyd1982least} is probably the most classical and widely used one for its effectiveness and efficiency.

Although $k$-means algorithm reaches remarkable performance under benign settings, however, it suffers from relatively poor performance when the data containing noise or outliers. This drawback relates to the latent connection between the Gaussian mixture model (GMM) \cite{Mclachlan1988GMM} and the thin-tailed property of the Gaussian distribution \cite{Peel2000Robust}. Several approaches have been proposed to improve the its robustness, which can be roughly divided into two main categories. Approaches in the first category conduct data clustering and outlier detection separately using a multi-stage method \cite{jiang2001two,hautamaki2005improving}, while those in the second category increase the robustness by alleviating the negative effects of outliers during the clustering process \cite{tseng2007penalized,chawla2013k,whang2015non}.  

In this paper, we revisit this problem from the aspect of the thin-tailed property of the Gaussian distribution. Instead of the Gaussian mixture distribution, we assume that the data is sampled from the $t$-mixture distribution, since $t$-distribution is a heavy-tailed generalization of the Gaussian distribution. Inspired by how the $k$-means algorithm is derived from the GMM, we derive a novel clustering method, dubbed $t$-$k$-means, from the $t$-mixture model (TMM) \cite{liu1995ml}. Accordingly, $t$-$k$-means enjoys the robust nature of TMM, and the efficiency of the $k$-means algorithm simultaneously. Besides, compared with the $k$-means algorithm, the proposed method is more stable from the aspect of the variance of multiple results. The robustness and stability of the proposed method are also analyzed from the aspect of the loss function and the expression of the clustering center, respectively. Moreover, we also provide a fast extension of the proposed method to further enhance its efficiency.

The main contributions of this paper are as follows:
\begin{itemize}
	\item We propose a novel clustering method, which is a robust and stable generalization of the $k$-means algorithm.
	\item We discuss the robustness and the stability of the proposed method theoretically from the aspect of the loss function and the expression of the clustering center, respectively.
	\item Extensive experiments are conducted, which empirically verify the effectiveness and efficiency of the proposed method.
\end{itemize}

\vspace{-1em}
\section{The Proposed Method}
\vspace{-0.6em}

\subsection{Preliminaries}\label{sec:gmm}
In this section, we briefly review the Gaussian mixture model (GMM) and its relation with $k$-means algorithm.

Given the dataset $\mathcal{D}=\{\pmb{x}_n\}_{n=1}^{N}$, where $\pmb{x}_n 
\in \mathbb{R}^p$ denotes a $p$-dim sample, GMM is a linear superposition of $K$-component Gaussian distribution \cite{Mclachlan1988GMM} with the probability density function given by
\begin{align}
\small
	\mathcal N(\pmb x|\pmb\pi, \pmb\mu,\pmb\Sigma)=\sum_{k=1}^K\pi_k\mathcal N_k(\pmb x|\pmb\mu_k,\pmb\Sigma_k),\label{eqn:gaussian}
\end{align}
where $\pmb\pi=\{\pi_k|k=1,\dots,K\}$, $\pmb\mu=\{\pmb\mu_k|k=1,\dots,K\}$, $\pmb\Sigma=\{\pmb\Sigma_k|k=1,\dots,K\}$, $\pi_k$ is non-negative and $\sum_{k=1}^K\pi_k=1$, $\pmb{\mu}_k \in \mathbb{R}^p$ and $\pmb{\Sigma}_k \in \Pi(p)$ are the mean vector and the covariance matrix of the $k$-th component, respectively. 

To estimate the parameters in GMM, the EM algorithm can be adopted \cite{bilmes1998gentle}. Let the complete-data of sample $\pmb x_n$ in the EM algorithm is denoted by $(\pmb x_n^\top, \pmb z_{n}^\top)^\top$, where the latent variable $z_{nk}=(\pmb z_n)_k\in\{1,0\}$ indicates whether $\pmb x_n$ belongs to the $k$-th component. In the M-step, the parameters $\pmb{\pi}, \pmb{\mu}$, and $\pmb{\Sigma}$ in GMM is updated by
\begin{align}\label{eqn:gmm_objective}
\small
\min_{\pmb\pi,\pmb\mu,\pmb\Sigma} -\ln \prod_{n=1}^{N}\prod_{k=1}^K[\pi_k \mathcal N_k(\pmb{x}_n|\pmb{\mu}_k, \pmb{\Sigma}_k)]^{r_{nk}},
\end{align}
where $r_{nk}$ is the expectation of $z_{nk}$. 

\newpage

Let $\pmb I$ denotes a $p$-dim identity matrix and $\alpha$ be a known parameter. Assuming that all the components share the same mixing coefficient and covariance matrix $\Sigma$ where $\Sigma = \alpha \pmb I$, $i.e.$, $\pi_k=\frac{1}{K},\pmb\Sigma_k=\alpha \pmb I$, $(k=1,\dots,K$). In this case,  Eq. \eqref{eqn:gmm_objective} becomes

\begin{equation}\label{eqn:kmeans_objective}
    \min_{\pmb\mu} \sum_{n=1}^{N} \sum_{k=1}^K r_{nk}(\pmb x_n-\pmb\mu_k)^\top(\pmb x_n-\pmb\mu_k).
\end{equation}

Eq. \eqref{eqn:kmeans_objective} is identical to the loss function of the $k$-means algorithm, $i.e.,$ $k$-means can be regarded as a special case of GMM \cite{mitchell1997machine}.

\vspace{-0.3em}
\subsection{$t$-$k$-means Algorithm}
To derive the $t$-$k$-means algorithm, we assume that the data is sampled from the $t$-mixture distribution with the probability density function given by  

\begin{align}
t(\pmb x|\pmb{\Psi})=\sum_{k=1}^K\pi_k t_k(\pmb x|\nu_k, \pmb{\mu}_k, \pmb{\Sigma}_k),
\end{align}
where $\pmb \Psi = \{\pmb\pi,\pmb\nu,\pmb\mu,\pmb\Sigma\}$ and $\pmb\nu=\{\nu_k|k=1,\dots,K\}$.

We now derive the $t$-$k$-means from the $t$-mixture model (TMM) also under the condition that $\Sigma = \alpha \pmb I, \pi_k=\frac{1}{K}$, and $\pmb\Sigma_k=\alpha \pmb I, (k=1,\dots,K$). Besides, we further assume that $\nu_k=\nu, (k=1,2,\dots,K)$ as suggested by Liu \textit{et al.} \cite{liu1995ml}. Those conditions are used to regulate the model complexity, so that the $t$-$k$-means algorithm can enjoy the running efficiency while preserving the robustness of TMM.

\vspace{0.4em}
\noindent \textbf{Complete-data Log Likelihood. } 
Similar to the complete-data vector $\pmb z$ in GMM, the complete-data vector in TMM is denoted by
\begin{align*}
	\pmb x_c=(\pmb x_1^\top,\dots,\pmb x_N^\top,\pmb z_1^\top,\dots,\pmb z_N^\top,u_1,\dots,u_N)^\top,
\end{align*}
where $\pmb z_1,\dots, \pmb z_N$ is also the cluster-related indicator and $u_1,\dots,u_N$ are the additional missing data \cite{liu1995ml}, such that
\begin{align}
	\begin{split}
	&\pmb x_n|u_n,z_{nk}=1\sim\mathcal N(\pmb\mu_k,\frac{\alpha\pmb I}{u_n}),\\
	&u_n|z_{nk}=1\sim \mathrm{gamma}(\frac{1}{2}\nu,\frac{1}{2}\nu).\label{eqn:XU}
	\end{split}	
\end{align}

\noindent Accordingly, the complete-data log likelihood is as follows:
\begin{equation}
    \ln L_c(\pmb\Psi|\pmb x,\pmb u,\pmb z)=\ln L_G(\nu|\pmb u,\pmb z)+\ln L_N(\pmb\mu,\alpha|\pmb x,\pmb u,\pmb z),
\end{equation}
where

\begin{align}
\small
	&\ln L_G(\nu|\pmb u,\pmb z)=\sum^K_{k=1}\sum^N_{n=1}z_{nk}\left\{-\ln\Gamma\left(\frac{1}{2}\nu\right)\right.\notag\\
	&\left.\qquad\qquad\quad+\frac{1}{2}\nu\ln\left(\frac{1}{2}\nu\right)+\frac{1}{2}\nu(\ln u_n-u_n)-\ln u_n\right\},\\
	&\ln L_N(\pmb\mu,\alpha|\pmb x,\pmb u,\pmb z)=\sum^K_{k=1}\sum^N_{n=1}z_{nk}\left\{-\dfrac{1}{2}p\ln(2\pi)\right.\notag\\
	&\qquad\qquad\qquad\;\left.-\frac{1}{2}\ln\frac{\alpha}{u_n}-\frac{u_n}{2\alpha}(\pmb x_n-\pmb\mu_k)^\top(\pmb x_n-\pmb\mu_k)\right\}.
\end{align}

\vspace{0.4em}
\noindent \textbf{EM-basedLog Likelihood Optimization. }
In the EM algorithm, the objective function in a new iteration is the current conditional expectation of the complete-data log likelihood, \textit{i.e.},
\begin{align}
	Q(\pmb\Psi^\star|\pmb\Psi)&=E(\ln L_c(\pmb\Psi|\pmb x,\pmb u,\pmb z))\notag\\&=Q_1(\nu^\star|\pmb\Psi)+Q_2(\pmb\mu^\star,\alpha^\star|\pmb\Psi),\label{eqn:Q}
\end{align}
where
\begin{align}
	&Q_1(\nu^\star|\pmb\Psi)=E(\ln L_G(\nu|\pmb u,\pmb z)),\\ &Q_2(\pmb\mu^\star,\alpha^\star|\pmb\Psi)=E(\ln L_N(\pmb\mu,\alpha|\pmb x,\pmb u,\pmb z)).
\end{align}
The parameters with superscript `$^\star$' will be estimated in the new iteration.

\vspace{0.3em}
\noindent \textbf{1) E-step}

\vspace{0.2em}
\noindent \textbf{(1) Estimate $E(z_{nk}|\pmb x_{n})$. } 

\begin{align}
	E(z_{nk}|\pmb x_{n})=\frac{t_k(\pmb x_n|\nu,\pmb\mu_k,\alpha\pmb I)}{\sum_{j=1}^Kt_j(\pmb x_n|\nu,\pmb\mu_j,\alpha\pmb I)}=\tau_{nk}.\label{eqn:tau}
\end{align}

\vspace{0.2em}
\noindent \textbf{(2) Estimate $E(u_n|\pmb x_n,\pmb z_n)$. } Since $\pmb x_n|u_n,z_{nk}=1\sim\mathcal N(\pmb\mu_k,\frac{\alpha\pmb I}{u_n})$, from the property of Gaussian distribution, we know $u_n(\pmb x_n-\pmb\mu_k)^\top(\pmb x_n-\pmb\mu_k)/\alpha$ follows $\chi^2_p$ distribution. Based on the property of gamma distribution, the likelihood of $u_n$ is
\begin{align}
	L(u_n|\pmb x_n)\propto\mathrm{gamma}\left(\frac{p}{2},\frac{(\pmb x_n-\pmb\mu_k)^\top(\pmb x_n-\pmb\mu_k)}{2\alpha}\right).\label{eqn:U}
\end{align}
According to Eq. \eqref{eqn:XU} and Eq. \eqref{eqn:U}, the posterior distribution of $u_n$ given $\pmb x_n,z_{nk}=1$ is
\begin{align}
	&u_n|\pmb x_n,z_{nk}=1\sim\mathrm{gamma}\left(\frac{\nu+p}{2},\right.\notag\\&\qquad\qquad\qquad\qquad\left.\frac{\nu+\frac{1}{\alpha}(\pmb x_n-\pmb\mu_k)^\top(\pmb x_n-\pmb\mu_k)}{2}\right).\label{eqn:u}
\end{align}
Based on Eq. \eqref{eqn:u}, we have
\begin{align}
	E(u_n|\pmb x_n,\pmb z_n)=\frac{\nu+p}{\nu+\frac{1}{\alpha}(\pmb x_n-\pmb\mu_k)^\top(\pmb x_n-\pmb\mu_k)}=u_{nk}.\label{eqn:u_nk}
\end{align}

\vspace{0.2em}
\noindent \textbf{(3) Estimate $E(\ln u_n|\pmb x_n,\pmb z_n)$.}

\begin{align}\label{eqn: un}
	E(\ln u_n|\pmb x_n,\pmb z_n)=\ln u_{nk}+\phi\left(\frac{\nu+p}{2}\right)-\ln\left(\frac{\nu+p}{2}\right).
\end{align}

Eq. \eqref{eqn: un} is obtained by applying the following Lemma. 

\vspace{0.3em}
\noindent \textbf{Lemma 1} \cite{liu1995ml}. \emph{If a random variable $R\sim\mathrm{gamma}(a,b)$, then $E(\ln R)=\phi(a)-\ln b$, where $\phi(a)=\{\partial\Gamma(a)/\partial a\}$.}

\vspace{0.7em}
\noindent \textbf{2) M-step}


\vspace{0.2em}
\noindent \textbf{(1) Estimate $\pmb\mu_k^\star$. } 
\begin{align}
	\frac{\partial Q_2(\pmb\mu^\star,\alpha^\star|\pmb\Psi)}{\partial \pmb\mu_k^\star}=0\implies\pmb\mu_k^\star=\frac{\sum^N_{n=1}\tau_{nk}u_{nk}\pmb x_n}{\sum^N_{n=1}\tau_{nk}u_{nk}}.\label{eqn:mu}
\end{align} 

\vspace{0.2em}
\noindent \textbf{(2) Estimate $\alpha^\star$. } Now we have
\begin{align}
\frac{\partial Q_2(\pmb\mu^\star,\alpha^\star|\pmb\Psi)}{\partial \alpha^\star}=0.
\end{align} 
Accordingly,
\begin{align}
	\alpha^\star=\frac{\sum_{k=1}^K\sum_{n=1}^N\tau_{nk}u_{nk}(\pmb x_n-\pmb\mu_k^\star)^\top(\pmb x_n-\pmb\mu_k^\star)}{p\sum_{k=1}^K\sum_{n=1}^N\tau_{nk}}.\label{eqn:alpha}
\end{align}

\begin{table*}[ht]
	\centering
	\caption{ARI of the clustering results on synthetic datasets.}
	\vspace{-0.2em}
	\begin{small}
	\begin{tabular}{l|ccccc}
		\toprule
		\multicolumn{1}{r}{} & A1    & A2    & A3    & S1    & S2  \\
		\midrule
		$k$-means & 0.804$\pm$0.068 & 0.807$\pm$0.056 & 0.829$\pm$0.039 & 0.844$\pm$0.059 & 0.826$\pm$0.057 \\
		$k$-means++ & 0.856$\pm$0.050 & 0.864$\pm$0.030 & 0.882$\pm$0.041 & 0.904$\pm$0.046 & 0.850$\pm$0.053 \\
		$k$-medoids & 0.775$\pm$0.081 & 0.783$\pm$0.057 & 0.792$\pm$0.039 & 0.817$\pm$0.056 & 0.803$\pm$0.070 \\
		$k$-medians & 0.760$\pm$0.060 & 0.780$\pm$0.060 & 0.780$\pm$0.040 & 0.810$\pm$0.070 & 0.780$\pm$0.070 \\
		GMM   & 0.088$\pm$0.013 & 0.052$\pm$0.008 & 0.035$\pm$0.012 & 0.127$\pm$0.009 & 0.122$\pm$0.002 \\
		TMM   & 0.483$\pm$0.189 & 0.295$\pm$0.153 & 0.264$\pm$0.110 & 0.409$\pm$0.185 & 0.483$\pm$0.143 \\
		\midrule
		$t$-$k$-means & 0.851$\pm$0.061 & 0.853$\pm$0.041 & 0.882$\pm$0.038 & 0.932$\pm$0.062 & 0.872$\pm$0.050 \\
		fast $t$-$k$-means & \underline{0.922$\pm$0.035} & \underline{0.928$\pm$0.025} & \underline{0.929$\pm$0.028} & \underline{0.986$\pm$0.000} & \underline{0.937$\pm$0.000} \\
		fast $t$-$k$-means++ & \textbf{0.954$\pm$0.045} & \textbf{0.948$\pm$0.021} & \textbf{0.945$\pm$0.020} & \textbf{0.986$\pm$0.000} & \textbf{0.936$\pm$0.000} \\
		\midrule
		\multicolumn{1}{r}{}     & S3    & S4    & Unbalance & dim32 & dim64 \\
		\midrule
		$k$-means & 0.639$\pm$0.039 & 0.584$\pm$0.026 & 0.589$\pm$0.306 & 0.650$\pm$0.081 & 0.639$\pm$0.091 \\
		$k$-means++ & 0.671$\pm$0.035 & 0.589$\pm$0.028 & \underline{0.909$\pm$0.078} & \underline{0.985$\pm$0.028} & \underline{0.995$\pm$0.018} \\
		$k$-medoids & 0.649$\pm$0.039 & 0.583$\pm$0.033 & 0.652$\pm$0.076 & 0.771$\pm$0.094 & 0.756$\pm$0.095 \\
		$k$-medians & 0.650$\pm$0.040 & 0.570$\pm$0.030 & 0.610$\pm$0.090 & 0.740$\pm$0.080 & 0.760$\pm$0.080 \\
		GMM   & 0.113$\pm$0.010 & 0.094$\pm$0.032 & 0.057$\pm$0.062 & 0.000$\pm$0.000   & 0.000$\pm$0.000 \\
		TMM  & 0.248$\pm$0.107 & 0.157$\pm$0.092 & 0.426$\pm$0.246 & 0.507$\pm$0.132 & 0.540$\pm$0.188 \\
		\midrule
		$t$-$k$-means & 0.699$\pm$0.028 & 0.612$\pm$0.011 & 0.829$\pm$0.169 & 0.968$\pm$0.065 & 0.938$\pm$0.102 \\
		fast $t$-$k$-means & \underline{0.718$\pm$0.018} & \underline{0.618$\pm$0.005} & 0.807$\pm$0.093 & 0.931$\pm$0.057 & 0.904$\pm$0.050 \\
		fast $t$-$k$-means++ & \textbf{0.726$\pm$0.011} & \textbf{0.623$\pm$0.000} & \textbf{0.931$\pm$0.076} & \textbf{0.997$\pm$0.015} & \textbf{1.000$\pm$0.000} \\
		\bottomrule
	\end{tabular}%
	\end{small}
	\label{tab:ari}%
	\vspace{-0.5em}
\end{table*}%

\vspace{0.2em}
\noindent \textbf{(3) Estimate $\nu^\star$. } The estimation of $\nu^\star$ is the solution of the equation
\begin{align}
	&-\phi\left(\dfrac{1}{2}\nu^\star\right)+\ln\left(\dfrac{1}{2}\nu^\star\right) + \eta =0, \label{eqn:nu_eq}
\end{align}
where
$\eta = 1+\frac{1}{K}\sum_{k=1}^K\dfrac{1}{\sum_{n=1}^N\tau_{nk}}\notag \sum_{n=1}^N\tau_{nk}(\ln u_{nk}-u_{nk})
	+\phi\left(\dfrac{\nu+p}{2}\right)-\ln\left(\dfrac{\nu+p}{2}\right)$ is a constant.

We apply Lemma 2 to solve Eq. \eqref{eqn:nu_eq}. Specifically, we have
\begin{align}
	\frac{1}{\nu^\star}+\epsilon(\nu^\star)+\eta\approx 0\implies\nu^\star\approx\frac{1}{-\eta-\epsilon(\nu^\star)},\label{eqn:nu}
\end{align}
where $\epsilon(\nu^\star) = \sum_{i=2}^\infty\frac{B_i}{i(\nu^\star)^i}$ is the error term.

Note that $\epsilon(\nu^\star)$ approximates to $0$ when $\nu\ge1$. Accordingly, we update $\nu^\star$ using $\frac{1}{-\eta}$.

\noindent \textbf{Lemma 2} \cite{abramowitz1964handbook}. \emph{$\phi(s)\approx \ln s-\sum_{i=1}^\infty\frac{B_i}{i(s)^i}$, where $B_i$ is the Bernoulli numbers of the second kind and $B_1=\frac{1}{2}$.}

\subsubsection{Fast $t$-$k$-means}
In TMM, if $\nu$ is unknown, the EM algorithm converges slowly \cite{liu1995ml}. As such, we fix $\nu$ as a constant as suggested in \cite{NIPS2009}. To further enhance the efficiency, we also apply $\alpha \to 0$ as suggested in \cite{Bishop2006Pattern}. With fixed $\nu$ and $\alpha \to 0$, we obtain a fast version of $t$-$k$-means, which is dubbed fast $t$-$k$-means.

\vspace{-0.6em}
\section{Robustness and Stability Analysis}
\vspace{-0.3em}

\subsection{Robustness Analysis}
The log likelihood of $t$-$k$-means is given by
\begin{align}
	\ln L(\pmb\Psi|\pmb x)&=\ln \prod_{n=1}^{N}\prod_{k=1}^K[ t_k(\pmb{x}_n|\nu, \pmb{\mu}_k, \alpha \pmb I)]^{z_{nk}}.\label{eqn:L_tkmeans}
\end{align}

\comment{
Given Eq. \eqref{eqn:L_tkmeans},we can rewrite the loss function of $t$-$k$-means as
\begin{align*}
	&J_{t\text{-}k\text{-means}}=-\sum_{n=1}^{N}\sum_{k=1}^K\tau_{nk}\ln \frac{\Gamma\left(\frac{\nu+p}{2}\right)}{\Gamma\left(\frac{\nu}{2}\right)\nu^{\frac{p}{2}}\pi^{\frac{p}{2}}\alpha^{\frac{1}{2}}}\cdot\\
	&\qquad\qquad\qquad\quad\left[1+\frac{1}{\nu\alpha}{(\pmb{x}_n-\pmb{\mu}_k)}^\top{(\pmb{x}_n-\pmb{\mu}_k)}\right]^{-\frac{\nu+p}{2}}.
\end{align*}0
}

Focusing on the term related to the data $\pmb x$, we have
\begin{align}
	&J_{t\text{-}k\text{-means}}(\pmb x,\pmb\mu)\propto\sum_{n=1}^{N}\sum_{k=1}^K\tau_{nk}\ln\left(1+\frac{1}{\nu\alpha}(\pmb{x}_n\right.\notag\\&\qquad\qquad\qquad\qquad\qquad\qquad-\pmb{\mu}_k)^\top(\pmb{x}_n-\pmb{\mu}_k)\bigg)\label{eqn:Jtkmeans}
\end{align}

According to  Eq. \eqref{eqn:Jtkmeans} and Eq. \eqref{eqn:kmeans_objective}, we know that $J_{t\text{-}k\text{-means}}$ is a $\log \ell^2$ loss function of $\pmb{x}_n$, while that of the $k$-means algorithm is a $\ell^2$ loss. Accordingly, $t$-$k$-means is more robust than $k$-means since its objective function is far less sensitive to the noise or outliers.

\subsection{Stability Analysis}
The randomness of the $k$-means and $t$-$k$-means methods is mainly involved in the selection of the initial clustering center. Once the initial clustering center is given, the clustering results of these methods are both fixed. In the $k$-means method, the update of clustering center is based only on the information of the sample in its cluster. However, according to Eq. (\ref{eqn:mu}), the update of the clustering center in $t$-$k$-means is determined by the information of all samples. In other words, no matter which sample is selected as the initial clustering center, the update of cluster centers still depends on all samples. This use of such `global information' significantly reduces the influence of the randomized clustering center in $t$-$k$-means algorithm, therefore it enjoys stronger stability.

\comment{
\begin{table}[ht]
\caption{Dataset description.}
\label{tab:data}
\vskip -0.1in
\begin{center}
\begin{small}
\begin{sc}
\begin{tabular}{lccc}
\toprule
Data set & Instances & Features & Clusters\\
\midrule
A1 & 3000 & 2 & 20\\
A2 & 5250 & 2 & 35\\
A3 & 7500 & 2 & 50\\
S1 & 5000 & 2 & 15\\
S2 & 5000 & 2 & 15\\
S3 & 5000 & 2 & 15\\
S4 & 5000 & 2 & 15\\
Unbalance & 6500 & 2 & 8\\
dim32 & 1024 & 32 & 16\\
dim64 & 1024 & 64 & 16\\
\cmidrule{1-4}
Iris & 150 & 4 & 3\\
Bezdekiris & 150 & 4 & 3\\
Seed & 210 & 7 & 3\\
Wine & 178 & 14 & 3\\
\bottomrule
\end{tabular}
\end{sc}
\end{small}
\end{center}
\vskip -0.15in
\end{table}
}

\section{Experiments}  

We conduct experiments on some synthetic \cite{ClusteringDatasets} and real-world datasets \cite{Lichman2013UCI}. In the experiments, the hyper-parameter $K$ is given by the selected datasets and the hyper-parameter $\nu$ in the fast $t$-$k$-means is set to $1$. The baselines include $k$-means \cite{lloyd1982least} , $k$-means++ \cite{Arthur2007kmeanspp}, $k$-medoids \cite{kaufman1987clustering}, $k$-medians \cite{arora1998kmedians}, GMM \cite{Mclachlan1988GMM}, and TMM \cite{liu1995ml}. When evaluating the performance of the model, adjusted rand index (ARI) \cite{Hubert1985Comparing}, clustering mean squared error (MSE) \cite{tan2006introduction}, and W/B (W: within-cluster sum of squares; B: between-cluster sum of squares) \cite{Kriegel2017WSS} are adopted. Besides, all experiment are repeated 100 times to reduce the effect of randomness. Among all methods, the one with the best performance is indicated in boldface and the value with underline denotes the second-best result. Besides, all methods are implemented with MATLAB on Intel(R) Core(TM) i7-7500U CPU running at 2.7GHz with 16 GB of RAM.

\begin{table*}[ht]
	\centering
	\caption{MSE and W/B of the clustering results on real-world datasets.}
	\vspace{-0.2em}
	\begin{small}
	\begin{tabular}{c|l|cccc}
		\toprule
		\multicolumn{1}{l|}{metrics} & methods & Bezdekiris & Iris  & Seed  & Wine \\
		\midrule
		\multirow{6}[3]{*}{MSE} & $k$-means & 0.201$\pm$0.033 & \underline{0.192$\pm$0.020} & \textbf{0.420$\pm$0.000} & 1.164$\pm$0.065 \\
		& $k$-means++ & 0.198$\pm$0.031 & 0.198$\pm$0.031 & \textbf{0.420$\pm$0.000} & \underline{1.154$\pm$0.000} \\
		& $k$-medoids & 0.205$\pm$0.037 & 0.222$\pm$0.048 & \underline{0.426$\pm$0.003} & 1.278$\pm$0.159 \\
		& $k$-medians & 0.215$\pm$0.045 & 0.226$\pm$0.051 & 0.434$\pm$0.053 & 1.258$\pm$0.138 \\
		& GMM  & 0.323$\pm$0.000 & 0.324$\pm$0.000 & 0.966$\pm$0.301 & 1.779$\pm$0.239 \\
		& TMM  & 0.425$\pm$0.238 & 0.293$\pm$0.106 & 0.606$\pm$0.121 & 1.612$\pm$0.209 \\
		\cmidrule{2-6}          & $t$-$k$-means & \textbf{0.186$\pm$0.000} & \textbf{0.187$\pm$0.000} & \textbf{0.420$\pm$0.000} & \underline{1.154$\pm$0.000} \\
		& fast $t$-$k$-means & \underline{0.187$\pm$0.000} & \textbf{0.187$\pm$0.000} & \textbf{0.420$\pm$0.000} & \textbf{1.153$\pm$0.000} \\
		& fast $t$-$k$-means++ & \underline{0.187$\pm$0.000} & \textbf{0.187$\pm$0.000} & \textbf{0.420$\pm$0.000} & \textbf{1.153$\pm$0.000} \\
		\midrule
		\multirow{6}[3]{*}{W/B} & $k$-means & 0.225$\pm$0.050 & \underline{0.213$\pm$0.030} & \textbf{0.329$\pm$0.000} & 0.989$\pm$0.159 \\
		& $k$-means++ & 0.222$\pm$0.046 & 0.222$\pm$0.046 & \textbf{0.329$\pm$0.000} & \textbf{0.966$\pm$0.001} \\
		& $k$-medoids & 0.242$\pm$0.049 & 0.263$\pm$0.063 & 0.358$\pm$0.032 & 1.117$\pm$0.224 \\
		& $k$-medians & 0.254$\pm$0.066 & 0.275$\pm$0.082 & 0.349$\pm$0.060 & 1.142$\pm$0.201 \\
		& GMM   & 0.418$\pm$0.000 & 0.419$\pm$0.000 & 2.349$\pm$3.241 & 3.280$\pm$2.102 \\
		& TMM   & 1.020$\pm$1.028 & 0.410$\pm$0.275 & 0.582$\pm$0.183 & 2.538$\pm$1.068 \\
		\cmidrule{2-6}          & $t$-$k$-means & \textbf{0.202$\pm$0.000} & \textbf{0.202$\pm$0.000} & \underline{0.333$\pm$0.000} & 1.015$\pm$0.000 \\
		& fast $t$-$k$-means & \underline{0.216$\pm$0.000} & 0.217$\pm$0.000 & \underline{0.333$\pm$0.000} & \underline{0.975$\pm$0.000} \\
		& fast $t$-$k$-means++ & \underline{0.216$\pm$0.000} & 0.217$\pm$0.000 & \underline{0.333$\pm$0.000} & \underline{0.975$\pm$0.000} \\
		\bottomrule
	\end{tabular}
	\end{small}
	\label{tab:real}
	\vspace{-0.5em}
\end{table*}

\vspace{-0.3em}
\subsection{Performance Evaluation}
\subsubsection{Performance on Synthetic Datasets}
In this section, we conduct experiments on synthetic datasets with labels, including S1~S4, A1~A3, Unbalance, dim32, and dim64 \cite{ClusteringDatasets}. 

As illustrated in Table~\ref{tab:ari}, the proposed $t$-$k$-means and fast $t$-$k$-means significantly outperform all other $k$-means variants, GMM, and TMM on all datasets. In particular, GMM, and TMM have relatively poor performance, probably because the mixture models demand heavy parameter estimation and are sensitive to the parameter initialization. Besides, the fast $t$-$k$-means++, which is obtained when fast $t$-$k$-means is initialized with $k$-means++ instead of random initialization, reaches the best performance on all 10 synthetic datasets. Moreover, the results of $t$-$k$-means-type method have a smaller standard deviation than that of other methods on all datasets, which empirically verifies the stability of the proposed  $t$-$k$-means.

\begin{table}[ht]
	\centering
	\caption{Running cost of different methods on Iris dataset.}
	\begin{small}
		\begin{tabular}{lcc}
			\toprule
			Methods & Iteration  & Total Time (sec)\\
			\midrule
		    $k$-means & 9.58 $\pm$ 1.67 & 0.0159 $\pm$ 0.0048 \\
			$k$-means++ & 8.50 $\pm$ 1.81 & 0.0156 $\pm$ 0.0047  \\
			$k$-medoids & 7.46 $\pm$ 1.20 & 0.0153 $\pm$ 0.0034  \\
			$k$-medians & 7.64 $\pm$ 1.38 & 0.0186 $\pm$ 0.0056  \\
			GMM   & 20.00$\pm$7.84 & 0.1462$\pm$0.0581 \\
			TMM   & 28.25$\pm$8.68 & 0.4136$\pm$0.1299 \\
			\cmidrule{1-3}          
			$t$-$k$-means & 29.76$\pm$5.82 & 0.1043$\pm$0.0228 \\
			fast $t$-$k$-means & 11.78$\pm$2.05 & 0.0183$\pm$0.0050 \\
			fast $t$-$k$-means++ & 10.50$\pm$2.87 & 0.0181$\pm$0.0074 \\
			\bottomrule
		\end{tabular}%
	\end{small}
	\label{tab:time}%
	\vspace{-0.5em}
\end{table}%

\subsubsection{Performance on Real-world Datasets}
To further verify the superiority of the proposed method, we evaluate our methods on $4$ real-world datasets without labels in this section. For Iris and Bezdekiris dataset, the labels are removed.

As demonstrated in Table \ref{tab:real}, $t$-$k$-means variants also reach the best performance in most cases on real-world datasets. Even when our methods could not perform the best, our methods still reach the second-best result and their performance is very close to the best result. Besides, the standard deviation of results performed by our methods is still smaller than that of other methods on all datasets, which further verifies the stability of the proposed $t$-$k$-means again.

\vspace{-1em}
\subsection{Running Efficiency}
\vspace{-0.5em}
In this section, we compare the running efficiency of different methods. As shown in Table~\ref{tab:time}, $t$-$k$-means require significantly less running time compared with TMM. Notably, the speed of fast $t$-$k$-means is on the same order of magnitude as that of the $k$-means algorithm, which verifies the efficiency of the proposed method.

\vspace{-0.5em}
\section{Conclusion}
\vspace{-0.5em}
In this paper, we propose a novel robust and stable $k$-means variant, the $t$-$k$-means, and its fast version based on the understanding of $k$-means, GMM, and TMM. We provide the detailed derivations for $t$-$k$-means, and compare its robustness and stability with $k$-means method from the aspect of the loss function and clustering center, respectively. Extensive experiments are conducted, which empirically demonstrate that the proposed method has better performance while preserving running efficiency.

\bibliographystyle{IEEEbib}
\bibliography{ref}

\end{document}